\documentclass[letterpaper, 10 pt, conference]{ieeeconf}  

\IEEEoverridecommandlockouts                              
\usepackage{amsmath} 
\usepackage{amssymb}  
\usepackage{amsfonts}
\usepackage{dsfont}
\usepackage{mathtools}
\usepackage{nicefrac}
\mathtoolsset{showonlyrefs}
\usepackage{xcolor}
\usepackage{tikz}
\usepackage[point]{fltpoint}
\usepackage{subcaption}

\usepackage{algorithm}
\usepackage{algorithmicx}
\usepackage{algpseudocode}

\usepackage{balance}

\include{math}

\def\Sec#1{Section~\ref{#1}}

\def\Fig#1{Fig.~\ref{#1}}

\def\Alg#1{Algorithm~\ref{#1}}

\algrenewcommand{\algorithmiccomment}[1]{$\triangleright$ \emph{#1}}

\def\DEF{\triangleq}

\definecolor{green}{rgb}{0,0.6,0}

\definecolor{red}{rgb}{0.6,0,0}

\usetikzlibrary{calc}

\graphicspath{{./}}

\title{\LARGE \bf
Bayesian Perceptron: Towards fully Bayesian Neural Networks
}

\author{Marco F. Huber
\thanks{This work was partially supported by the Ministry of Economic Affairs of the state Baden-W\"urttemberg (Center for Cyber Cognitive Intelligence (CCI) -- Grant No. 017-192996 and KI-Fortschrittszentrum ``Lernende Systeme'' -- Grant No. 036-170017).}
\thanks{M.F.~Huber is with the Institute of Industrial Manufacturing and Management~IFF, University of Stuttgart, 70569 Stuttgart, Germany. He further is with the Center for Cyber Cognitive Intelligence (CCI) as well as with the Signal and Image Processing Department, Fraunhofer Institute for Manufacturing Engineering and Automation IPA, 70569 Stuttgart, Germany.
\newline Email: {\tt\small marco.huber@ieee.org}\newline ORCID: 0000-0002-8250-2092}%
}

\begin{document}

\maketitle
\thispagestyle{empty}
\pagestyle{empty}

\begin{abstract}
Artificial neural networks (NNs) have become the de facto standard in machine learning. They allow learning highly nonlinear transformations in a plethora of applications. However, NNs usually only provide point estimates without systematically quantifying corresponding uncertainties. In this paper a novel approach towards fully Bayesian NNs is proposed, where training and predictions of a perceptron are performed within the Bayesian inference framework in closed-form. The weights and the predictions of the perceptron are considered Gaussian random variables. Analytical expressions for predicting the perceptron's output and for learning the weights are provided for commonly used activation functions like sigmoid or ReLU. This approach requires no computationally expensive gradient calculations and further allows sequential learning.
\end{abstract}

\section{Introduction}
\label{sec:introduction}
Deep artificial neural networks (NNs) are the driver behind many breakthroughs we have seen in applications like computer vision \cite{He_ICCV2017}, robotics \cite{tm2019}, or games \cite{Silver2016}. This is mainly due to their ability of accurately learning highly nonlinear functions from data in an end-to-end manner. Despite this success, high prediction accuracy is not sufficient in safety-critical applications like autonomous driving \cite{Loquercio2020} or human-robot-collaboration \cite{el-shamouty_icra2020}, where the robustness of the predictions and uncertainty quantification are additional requirements. Violating these requirements gives rise to problems like adversarial attacks \cite{Akhtar2018, Ranjan2019}, where NN are confused with specially designed data and patterns. 

One way to overcome the limitations of standard NNs is to combine them with Bayesian inference. This combination known as \emph{Bayesian NN} allows benefiting from the representational power of NNs on the one hand and from the principled parameter estimation of Bayesian inference on the other hand. First approaches to Bayesian NN date back to the early 1990s, where David MacKay demonstrated in \cite{MacKay1992} the various benefits of using Bayesian inference techniques for training NNs. Exact Bayesian inference for estimating the weights of an NN, however, is intractable due to the nonlinear nature and the number of parameters to be estimated. Thus, approximations are inevitable. A commonly used approximation technique is based on variational inference \cite{Graves2011, Blundell2015, Foong2019}, where the true posterior probability distribution of the NN's weights is approximated by means of a parametric distribution, typically a Gaussian. Estimating the parameters of this approximate distribution cannot be performed in closed-form in general. Instead, Monte Carlo sampling and gradient descent are usually employed, which makes Bayesian NN training computationally expansive and causes problems of controlling the high variance of the Monte Carlo gradient estimates.

An alternative to variational inference for approximating the Bayesian posterior over the weights is dropout. It is shown in \cite{Gal_ICML2016} that dropout allows uncertainty quantification and corresponds to an approximation of the variational distribution. In \cite{Puskorius2001} the training of a Bayesian NN is treated as a Kalman filtering problem, but as with variational inference, gradient descent is necessary to calculate the filtering matrices.

\begin{figure}[t]%
\centering
\begin{tikzpicture}
				\begin{scope}
					\clip (0, 0) circle (1.2cm);
					\draw[thick, fill=white] (0,0) circle (1.2cm);
					\filldraw[lightgray] (.9,-1.2) rectangle (1.2,1.2);
					\draw[thick,] (.9,-1.2) -- (.9,1.2);
				\end{scope}
				\draw[thick] (0,0) circle (1.2cm) node (nucleus) {};
				\node at ($(nucleus)+(-.15,0)$) {\small $\sum_i\hspace{-.5mm} w_i \hspace{-.5mm}\cdot\hspace{-.5mm} x_i\hspace{-.5mm} + \hspace{-.5mm} w_0$};
				\node at (1.05,0) (act) {$f$};
				
				\draw[thick, latex-] ($(nucleus)+(-1.1,.5)$) -- +(-1.3,.6) node[midway, above=-1mm] (w) {\rotatebox{-22}{\scalebox{.75}{$w_1$}}} node[at end] (w1) {};
				\draw[thick] ($(w1)+(-.15,.12)$) circle (2mm) node {\scalebox{.75}{$x_1$}};
				\draw[thick, latex-] ($(nucleus)+(-1.2,0)$) -- +(-1.3,0) node[midway, above=-.5mm] {\scalebox{.75}{$w_2$}} node[at end] (w2) {};
				\draw[thick] ($(w2)+(-.19,0)$) circle (2mm) node {\scalebox{.75}{$x_2$}};
				\draw[thick, latex-] ($(nucleus)+(-1.1,-.5)$) -- +(-1.3,-.6) node[midway, below=-1mm] {\rotatebox{25}{\scalebox{.75}{$w_d$}}} node[at end] (w3) {};
				\draw[thick] ($(w3)+(-.15,-.12)$) circle (2mm) node {\scalebox{.75}{$x_d$}};
				\node at (-2.67,-.48) {\rotatebox{7}{\scalebox{1.4}{$\vdots$}}};
				\draw[thick, latex-] ($(nucleus)+(0,-1.2)$) -- +(0,-.5) node[below] (b) {$w_0$};
				
				\draw[thick] ($(nucleus)+(1.2,0)$) -- +(.75,0) node[at end] (axon) {};
				\draw[thick, -latex] (axon.center) -- +(.8,1);
				\draw[thick, -latex] (axon.center) -- +(.8,.5);
				\draw[thick, -latex] (axon.center) -- +(.8,-.5);
				\draw[thick, -latex] (axon.center) -- +(.8,-1);
				
				\node at (-1.5,-1.9) (in) {Input};
				\filldraw[thick] (in.170) -- ($(w3)+(0,-.25)$) circle (.4mm);
				\node at (-.8,-1.4) (bi) {Bias};
				\filldraw[thick] (bi.320) -- ($(b)+(-.3,.1)$) circle (.4mm);
				\node at (-.5,1.4) (we) {Weights};
				\filldraw[thick] (we.west) -- ($(w)+(.1,.2)$) circle (.4mm);
				\node[text width=1.8cm, align=left] at (2.0,-1.7) (a) {Activation function};
				\filldraw[thick] (a.140) -- ($(act)+(0,-.3)$) circle (.4mm);
	\end{tikzpicture}
\caption{Perceptron, building block of artificial NNs.}%
\label{fig:perceptron}%
\end{figure}
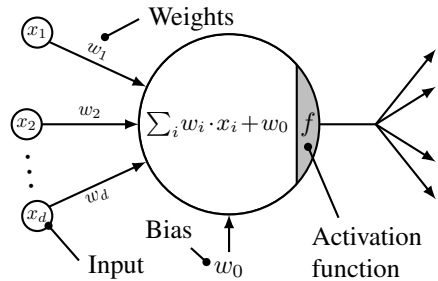

In this paper, the focus is on the core building block of Bayesian NNs, which is a probabilistic version of the perceptron/neuron (cf. \Fig{fig:perceptron} for a standard perceptron). Based on the common assumption that the weight distribution can be represented by means of a Gaussian, a novel approach for estimating the parameters of this distribution, i.e., its mean and covariance, is proposed. For this purpose the following contributions are made:

\begin{itemize}
	\item Closed-form propagation of the parameters of the perceptron's output distribution for commonly used activation functions like sigmoid or ReLU.
	\item Closed-form estimation of the parameters of the weight distribution for given training data without the need of gradient descent or Monte Carlo sampling. Instead, the Bayesian inference paradigm is strictly adhered to.
	\item Training data can be processed sequentially, while common Bayesian NN approaches require batch processing.
\end{itemize}

This approach---named \emph{Bayesian perceptron} (BP) in the following---is a first step towards building and training also deep NNs in a fully Bayesian manner. 

The paper is structured as follows: In the next section a problem statement is given. \Sec{sec:forward} defines the forward pass of the BP for estimating the output moments in closed form, while \Sec{sec:backward} gives the closed-form and sequential procedure for training the weights. The novel approach is validated in \Sec{sec:validation} with synthetic data. The paper closes with conclusions and an outlook to future work.

\section{Problem Formulation}
\label{sec:problem}
In this paper supervised machine learning problems are considered. For training purposes a training dataset $\SD = \{\vx_i, y_i\}_{i=1}^n$ comprising $n$ i.i.d. training instances $(\vx_i, y_i)$, with inputs/features $\vx_i = \[x_{i,1}\ \ldots x_{i,d}\]\T \in \NewR^d$ and outputs $y_i \in \NewR$, is given. In case of binary classification problems 
the outputs $y_i$ take values from the discrete set $\{0,1\}$.

The core building block of commonly employed NNs is the so-called \emph{perceptron} and variations of it. A perceptron as depicted in \Fig{fig:perceptron} learns a nonlinear transformation from an input $\vx$ to the scalar output $y$ by means of
\begin{align}
	\label{eq:problem_linear}
	a &= \vx\T\cdot \vw + w_0~,\\
	y &= f(a)~,
	\label{eq:problem_activation}
\end{align}
where $\vw = \[w_1 \ldots w_d\]\T \in \NewR^d$ comprises the weights, $w_0$ is the so-called bias, and $f(.)$ is the nonlinear activation function. To simplify notation, the convention of including the bias into the weight vector is employed in the following. That is, \eqref{eq:problem_linear} becomes $a = \vx\T\cdot \vw$ with $\vx = \[1\ x_1\ x_2\ \ldots\ x_d\]\T$ and $\vw = [w_0\ w_1\ \ldots w_d]\T$.

The originally proposed perceptron \cite{Rosenblatt1958} utilizes the Heaviside step function. In modern deep NNs, other activation functions proved to be more suitable. Thus, in this paper two main classes of commonly utilized activation functions are considered: (i) sigmoidal functions and (ii) piece-wise linear (pwl) functions. The class of sigmoidal (s-shaped) activations comprises 
\begin{align}
	\label{eq:problem_sigmoid}
	f(a) &= s(a) \DEF \frac{1}{1 + \text{e}^{-a}} & \text{(sigmoid)}~, \\
	\label{eq:problem_tanh}
	f(a) &= \tanh(a) = 2\cdot s(a) + 1& \text{(hyperbolic tangent)}~.
\end{align}
Due to the linear relation between sigmoid and hyperbolic tangent, w.l.o.g. only the sigmoid function is considered in the following. Piece-wise linear activations are given by\footnote{Please note that this paper can be easily extended to pwl activations with more than two pwl elements. We restrict ourselves to two elements in order to cover the most commonly used activations and to keep the notation uncluttered.}
\begin{equation}
	f(a) = \max(\alpha \cdot a, \beta\cdot a)~,
	\label{eq:problem_pwl}
\end{equation}
with $\alpha \in [0,1]$, $\beta \ge 0$, and $\alpha \le \beta$. This definition comprises the important special cases \emph{rectified linear unit (ReLU)} \cite{Hahnloser_Nature2000} for $\alpha = 0$, $\beta=1$, \emph{leaky ReLU} for $\beta=1$, and \emph{linear} activation for $\alpha=\beta=1$.

In standard NNs, the weights of the perceptrons are deterministic values or point estimates. Bayesian NNs instead use weights that are assigned a probability distribution. In this paper it is assumed that the weight vector $\rvw \sim \Gauss\big(\vmu^w, \C^w\big)$ in \eqref{eq:problem_linear} is Gaussian\footnote{Random variables are denoted by lower-case bold letters.} with mean vector $\vmu^w$ and covariance matrix $\C^w$. In doing so, a perceptron becomes a probabilistic model with $p(y|\SD)$ being the probability distribution of the output given the training data. 
When dealing with such a probabilistic model, two key tasks have to be performed: 
(i) \emph{prediction}, i.e., estimating the probability density function (pdf) $p(y|\vx, \SD)$ of the output~$\y$ given a so far unseen input~$\vx$ and 
(ii) \emph{training}, i.e., estimating the pdf $p(\vw|\SD)$ and its parameters $\vmu^w$, $\C^w$, respectively, given the training data $\SD$ and a prior pdf $p(\vw)$ of the weights~$\rvw$. For both tasks closed-form solutions for calculating the parameters of the corresponding pdfs are derived.

\section{Bayesian Perceptron: Forward Pass}
\label{sec:forward}
In this section the first task of predicting the output distribution of a BP is solved. 
For this purpose we assume a given test input $\vx$ that is passed forward through the perceptron allowing for the calculation of the predictive pdf~$p(y|\vx,\SD)$.

\subsection{Predictive Distribution}
\label{sec:forward_output}
Due to the nonlinearity introduced by the activity function $f(.)$ an analytical calculation of the exact predictive pdf is only possible in some special cases. 
Instead, it must be approximated, applying the usual Bayesian NN assumption that the predictive pdf can be approximated well by means of a parametric distribution, particularly a Gaussian, i.e., $p(y|\vx, \SD) \approx \Gauss(y; \mu_y, \sigma_y^2)$ with mean $\mu_y$ and variance $\sigma_y^2$. In doing so, calculating the predictive pdf boils down to calculating its parameters, i.e.,
\begin{align}
	\label{eq:muy}
	\mu_y &= \E\{y\} = \E\{f(a)\}~,
	\\
	\label{eq:sigmay}
	\sigma_y^2 &= \E\big\{(y-\mu_y)^2\big\} = \E\big\{f(a)^2\big\} - \mu_y^2~.
\end{align}
To solve the involved expected values, it can be exploited that for $\rvw$ being Gaussian also $\a\sim\Gauss(\mu_a, \sigma_a^2)$ is a Gaussian random variable with mean and variance according to
\begin{align}
	\label{eq:mua}
	\mu_a &= \vx\T \cdot \vmu^w~,
	\\
	\label{eq:sigmaa}
	\sigma_a^2 &= \vx\T\cdot \C^w \cdot \vx~,
\end{align}
respectively, which follows from the linearity of \eqref{eq:problem_linear} allowing applying the well-known Kalman prediction step \cite{Kalman1960, Habil2015}. 

\subsection{Sigmoid Case}
\label{sec:forward_sigmoid}
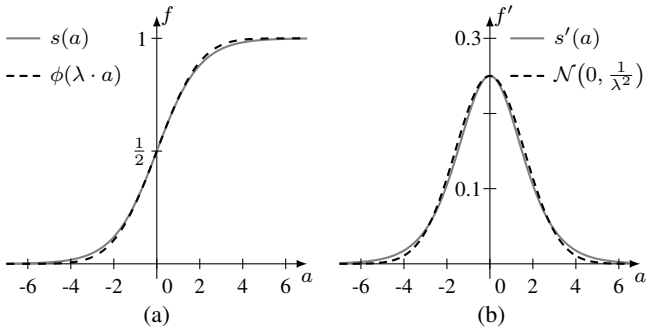
\begin{figure}[t]%
\centering
\begin{tikzpicture}
  \tikzstyle{axes}=[]

  \fpAdd{\xmin}{-7}{0}\fpAdd{\xmax}{7}{0} 
  \fpAdd{\ymax}{1}{0}   
  \fpAdd{\xsize}{4}{0}
  \fpAdd{\ysize}{3}{0}
   
  \fpSub{\dx}{\xmax}{\xmin}
  \fpDiv{\xscale}{\xsize}{\dx}
  \fpDiv{\yscale}{\ysize}{\ymax}
  
  \begin{scope}[style=axes, >=latex]
    \draw[->] (\xmin*\xscale,0) -- (\xmax*\xscale,0) node[right,below]  {\footnotesize$a$};
    \draw[->] (0,-.2) -- (0,\ymax*\yscale+.3) node[right=-2pt]  {\footnotesize$f$};
  \end{scope}
  
  \begin{scope}[style=axes, domain=\xmin:\xmax, xscale=\xscale, yscale=\yscale]
  	\draw[thick, gray] plot[smooth] file{sigmoid.dat};
  	\draw[thick, dashed] plot[smooth] file{probit.dat};
  \end{scope}
  
 	\draw (-6*\xscale,-2pt) -- (-6*\xscale,2pt) node[right,below=4pt] {\footnotesize-6};
	\draw (-4*\xscale,-2pt) -- (-4*\xscale,2pt) node[right,below=4pt] {\footnotesize-4};
	\draw (-2*\xscale,-2pt) -- (-2*\xscale,2pt) node[right,below=4pt] {\footnotesize-2};
 	\draw (0,-2pt) -- (0,2pt) node[right=4pt,below=4pt] {\footnotesize0};
 	\draw (2*\xscale,-2pt) -- (2*\xscale,2pt) node[right,below=4pt] {\footnotesize2};
	\draw (4*\xscale,-2pt) -- (4*\xscale,2pt) node[right,below=4pt] {\footnotesize4};
	\draw (6*\xscale,-2pt) -- (6*\xscale,2pt) node[right,below=4pt] {\footnotesize6};
  \draw (-2pt,.5*\yscale) -- (2pt,.5*\yscale) node[left=2pt] {\footnotesize$\tfrac{1}{2}$};
  \draw (-2pt,1*\yscale) -- (2pt,1*\yscale) node[left=2pt] {\footnotesize1};
         
  \begin{scope}[thick]
  	\draw[thick, gray] (-2,3) -- (-1.55,3) node[right] {\color{black}\footnotesize$s(a)$};
  	\draw[thick, densely dashed] (-2,2.5) -- (-1.55,2.5) node[right] {\footnotesize$\phi(\lambda \cdot a)$};
  \end{scope}
	
	\node at (0,-.7) {\small (a)};
	

	\fpAdd{\xmin}{8.5}{0}\fpAdd{\xmax}{22.5}{0} 
  \fpAdd{\ymax}{.3}{0}   
	\fpAdd{\ysize}{3}{0}
	\fpDiv{\yscale}{\ysize}{\ymax}
	
  \begin{scope}[style=axes, >=latex]
    \draw[->] (\xmin*\xscale,0) -- (\xmax*\xscale,0) node[right,below]  {\footnotesize$a$};
    \draw[->] (15.5*\xscale,-.2) -- (15.5*\xscale,\ymax*\yscale+.3) node[right=-2pt]  {\footnotesize$f'$};
  \end{scope}
	
  \begin{scope}[style=axes, domain=\xmin:\xmax, xscale=\xscale, yscale=\yscale]
  	\draw[thick, gray] plot[smooth] file{dsigmoid.dat};
  	\draw[thick, densely dashed] plot[smooth] file{gauss.dat};
  \end{scope}
	
 	\draw (9.5*\xscale,-2pt) -- (9.5*\xscale,2pt) node[right,below=4pt] {\footnotesize-6};
	\draw (11.5*\xscale,-2pt) -- (11.5*\xscale,2pt) node[right,below=4pt] {\footnotesize-4};
	\draw (13.5*\xscale,-2pt) -- (13.5*\xscale,2pt) node[right,below=4pt] {\footnotesize-2};
 	\draw (15.5*\xscale,-2pt) -- (15.5*\xscale,2pt) node[right=4pt,below=4pt] {\footnotesize0};
 	\draw (17.5*\xscale,-2pt) -- (17.5*\xscale,2pt) node[right,below=4pt] {\footnotesize2};
	\draw (19.5*\xscale,-2pt) -- (19.5*\xscale,2pt) node[right,below=4pt] {\footnotesize4};
	\draw (21.5*\xscale,-2pt) -- (21.5*\xscale,2pt) node[right,below=4pt] {\footnotesize6};
  \draw (15.2*\xscale,.1*\yscale) -- (15.8*\xscale,.1*\yscale) node[left=2pt] {\footnotesize0.1};
  \draw (15.2*\xscale,.2*\yscale) -- (15.8*\xscale,.2*\yscale) node[left=2pt] {};
	\draw (15.2*\xscale,.3*\yscale) -- (15.8*\xscale,.3*\yscale) node[left=2pt] {\footnotesize0.3};
	
  \begin{scope}[thick]
  	\draw[thick, gray] (4.7,3) -- (5.15,3) node[right] {\color{black}\footnotesize$s'(a)$};
  	\draw[thick, densely dashed] (4.7,2.5) -- (5.15,2.5) node[right] {\footnotesize$\Gauss\big(0, \tfrac{1}{\lambda^2}\big)$};
  \end{scope}
	
	\node at (4.45,-.7) {\small (b)};
\end{tikzpicture}
\vspace{-3mm}
\caption{Approximations of the sigmoid function (a) and its derivative (b) by means of the probit function and a Gaussian pdf, respectively, for $\lambda = \sqrt{\nicefrac{\pi}{8}}$\ .}%
\label{fig:sigmoid}%
\end{figure}

For sigmoidal activation functions it is well known that both expected values in \eqref{eq:muy} and \eqref{eq:sigmay} cannot be evaluated in closed form. For the mean $\mu_y$ a close approximation can be found, if the so-called \emph{probit} function $\phi(a) = \nicefrac{1}{2}\cdot(1+\erf(\nicefrac{a}{\sqrt{2}}))$ is substituted for the sigmoid with $\erf(.)$ being the Gaussian error function. The probit function is also s-shaped and the cumulative distribution function of the standard Gaussian pdf. Scaling the input of the probit function by $\lambda > 0$ leads~to
\begin{align}
	\mu_y 
	= \E\{s(a)\}
	&= \int_\NewR s(a)\cdot \Gauss\hspace{-.7mm}\(a; \mu_a, \sigma_a^2\) \dd a \\
	&\approx \int_\NewR \phi(\lambda \cdot a) \cdot \Gauss\hspace{-.7mm}\(a; \mu_a, \sigma_a^2\) \dd a \\
	&\hspace{-.5mm}\stackrel{(a)}{=} \phi\hspace{-.5mm}\(\tfrac{\lambda\cdot\mu_a}{t}\) \stackrel{(b)}{\approx} s\hspace{-.5mm}\(\tfrac{\mu_a}{t}\)~,
	\label{eq:muy_sigmoid}
\end{align}
with $t \DEF \sqrt{1+\lambda^2\cdot \sigma_a^2}$, where $(a)$ is a well-known solution to this integral (cf.~\cite{Murphy2012, Kristiadi2020}) and $(b)$ follows from re-substituting the scaled probit function with the sigmoid function. As stated in~\cite{Murphy2012, Kristiadi2020} a particularly well-fitting approximation of the sigmoid by the probit function is given for $\lambda = \sqrt{\nicefrac{\pi}{8}}$ as depicted in \Fig{fig:sigmoid}(a).

For the approximation \eqref{eq:muy_sigmoid} the following properties hold: (i) It is limited to the interval $[0,1]$ and thus, cannot deviate arbitrarily from the true value, which itself is bound to the same interval. (ii) For (almost) deterministic weights, i.e., $\sigma_a \rightarrow 0$, it approaches $s(\mu_a)$. Thus, the approximation correctly captures the deterministic special case. (iii) For increasing uncertainty, i.e., $\sigma_a \rightarrow \infty$, it approaches $s(0) = \nicefrac{1}{2}$. As $\mu_y = \text{Prob}(y=1|\vx, \SD)$ (cf. \cite{Kristiadi2020}), this limit reflects that the perceptron is ``indifferent'' if there is high uncertainty, which is as expected.

To calculate the variance \eqref{eq:sigmay} of $\y$, for $f(a) = s(a)$ it holds
\begin{align}
	\sigma_y^2 
	&= \E\big\{s(a)^2\big\} - \mu_y^2 \\
	&= \E\big\{s(a)^2 + s(a) - s(a) \big\} -\mu_y^2 \\
	&\hspace{-.5mm}\stackrel{(c)}{=} \E\big\{s(a) - \underbrace{s(a)\cdot(1-s(a))}_{= s'(a)}\big\} -\mu_y^2 \\
	&= \mu_y - \mu_y^2 - \E\big\{s'(a)\big\}~,
	\label{eq:sigmay_derivative}
\end{align}
where in $(c)$ the fact is exploited that the derivative of the sigmoid function can be described by means of sigmoids. The remaining expected value in \eqref{eq:sigmay_derivative} has no analytical solution. However, by again substituting the (scaled) probit function for the sigmoid we can make use of the fact that the derivative of the probit function is the Gaussian pdf $\Gauss(a; 0, \nicefrac{1}{\lambda^2})$. This yields a close approximation of $s'(a)$ as depicted in \Fig{fig:sigmoid}(b). Hence, it follows for the expected value in \eqref{eq:sigmay_derivative}
\begin{align}
	\E\big\{s'(a)\big\} 
	&\approx \E\big\{\Gauss\hspace{-.7mm}\(a; 0, \tfrac{1}{\lambda^2}\)\big\}\\ 
	&= \int_\NewR \Gauss\hspace{-.7mm}\(a; 0, \tfrac{1}{\lambda^2}\)\cdot \Gauss\hspace{-.7mm}\(a; \mu_a, \sigma_a^2\)\dd a \\
	&\hspace{-.5mm}\stackrel{(d)}{=} \Gauss\hspace{-.7mm}\(0; \mu_a, \tfrac{1}{\lambda^2} + \sigma_a^2\) \\
	&= \tfrac{1}{t}\cdot \Gauss\hspace{-.7mm}\(\tfrac{\mu_a}{t}; 0, \tfrac{1}{\lambda^2}\) \stackrel{(e)}{\approx} \tfrac{1}{t}\cdot s'\hspace{-.7mm}\(\tfrac{\mu_a}{t}\)~,
	\label{eq:expected_derivative}
\end{align}
where $(d)$ results from knowing that the product of two Gaussian pdfs is an unnormalized Gaussian for which the integral can be solved analytically. $(e)$ results from re-substituting the Gaussian with the sigmoid derivative. Plugging \eqref{eq:expected_derivative} in~\eqref{eq:sigmay_derivative} yields the desired (close) approximation of the variance according to
\begin{align}
	\sigma_y^2 \approx \mu_y\cdot(1-\mu_y)\cdot(1 - \tfrac{1}{t})~,
	\label{eq:sigmay_sigmoidapprox}
\end{align}
by means of exploiting that $s'(a) = s(a)\cdot (1-s(a))$ and $\mu_y \approx s(\nicefrac{\mu_a}{t})$ due to \eqref{eq:muy_sigmoid}.

For the approximation in \eqref{eq:sigmay_sigmoidapprox} it  can be shown that it is bounded to the interval $[0, \nicefrac{1}{4}]$, where the lower bound follows from $\sigma_a \rightarrow 0$ (deterministic case) and the upper bound from $\sigma_a \rightarrow \infty$ (high uncertainty).

\subsection{Piece-wise Linear Case}
\label{sec:forward_pwl}
While the sigmoid activation function requires approximations for calculating the both parameters \eqref{eq:muy} and \eqref{eq:sigmay} of the predictive pdf of the output $\y$, the involved integrals can be solved analytically exactly for pwl activations \eqref{eq:problem_pwl} up to the Gaussian error function. In case of the predicitive mean, with $p_x(a) \DEF p(a|\vx, \SD) = \Gauss(a; \mu_a, \sigma_a^2)$ it holds that
\begin{align}
	\mu_y 
	&= \E\{f(a)\} \\
	&= \int_\NewR \max(\alpha\cdot a, \beta\cdot a) \cdot p_x(a)\dd a \\
	&\hspace{-.5mm}\stackrel{(f)}{=} \alpha\cdot \int_{-\infty}^0 a \cdot p_x(a)\dd a + \beta\cdot \int_0^\infty a \cdot  p_x(a)\dd a \\
	&= \alpha\cdot \underbrace{\int_\NewR a \cdot p(a|\vx)\dd a}_{= \mu_a\DEF \E_1} + (\beta-\alpha)\cdot \int_0^\infty a \cdot  p_x(a)\dd a \\
	&= \alpha\cdot \E_1 +\, (\beta-\alpha)\cdot \(\E_1\cdot\, \phi\(\tfrac{\mu_a}{\sigma_a}\) + p_a\),\quad~
	\label{eq:my_pwl}
\end{align}
where $(f)$ follows from $\alpha \le \beta$, $\E_1$ is the first raw Gaussian moment, and $p_a\DEF\sigma_a^2\cdot p_x(0)$. Similarly, for the predictive variance it follows
\begin{align}
	\sigma_y^2
	&= \E\{f(a)^2\} - \mu_y^2 \\
	&= \int_\NewR \(\max(\alpha\cdot a, \beta\cdot a)\)^2 \cdot p_x(a)\dd a - \mu_y^2\\
	&= \alpha^2\cdot \hspace{-.4mm}\int_{-\infty}^0\hspace{-.4mm} a^2 \cdot p_x(a)\dd a + \beta^2\cdot\hspace{-.4mm} \int_0^\infty\hspace{-.4mm} a^2 \cdot  p_x(a)\dd a - \mu_y^2 \\
	&= \alpha^2\cdot \underbrace{\int_\NewR a^2\cdot p_x(a) \dd a}_{= \mu_a^2 + \sigma_a^2\DEF \E_2} +\, c \cdot \int_0^{\infty} a^2 \cdot p_x(a)\dd a - \mu_y^2 \\
	&= \alpha^2\cdot\E_2 +\, c\cdot\Big(\E_2\cdot\,\phi\(\tfrac{\mu_a}{\sigma_a}\) + \mu_a\cdot p_a\Big) - \mu_y^2~,
	\label{eq:sigmay_pwl}
\end{align}
with $c \DEF \(\beta^2-\alpha^2\)$ and $\E_2$ being the second raw Gaussian moment.

It is important to note that the predictive distribution $p(y|\vx, \SD)$ is approximated by a Gaussian $\Gauss(y; \mu_y, \sigma_y^2)$ with the exact predictive
mean \eqref{eq:my_pwl} and the exact predictive variance \eqref{eq:sigmay_pwl}. This approximation is known as \emph{moment matching} in general and thus, this is a very efficient form of assumed density filtering, which has previously been introduced by~\cite{Maybeck1979} in the area of Bayesian filtering. For the sigmoid case both moments of the predictive distribution are calculated almost exactly by \eqref{eq:muy_sigmoid} and \eqref{eq:sigmay_sigmoidapprox} in a computationally lightweight fashion.

\section{Bayesian Perceptron: Backward Pass}
\label{sec:backward}
While the previous section was concerned with inferring the output $\y$ given an arbitrary input $\vx$, which corresponds to a forward pass through the BP, this section deals with the backward pass. Here, the task is to update the weights $\rvw$ given training data $\SD$. Thanks to the common assumption of i.i.d. training instances, updating the weights can be performed sequentially, i.e., each training instance $(\vx_i, y_i) \in \SD$ is processed sequentially. In doing so, there is no need for iterative batch processing being common in training NNs. 

Given the prior distribution $p_{i-1}(\vw)~\DEF~p(\vw|\SD_{i-1})~=$ $\Gauss(\vw, \vmu_{i-1}^w, \C_{i-1}^w)$ resulting from processing all training instances $\SD_{i-1} \DEF \{(\vx_1, y_1)\ \ldots\ (\vx_{i-1}, y_{i-1})\}$, updating the perceptron's weights $\rvw$ by means of the so far unseen $i$-th \linebreak training instance $(\vx_i, y_i)$ corresponds to calculating the posterior distribution $p_i(\vw)$ due to the Bayesian nature of the considered perceptron. Again, the posterior distribution cannot be calculated analytically in general. A Gaussian distribution is used for approximating the true posterior, which captures the posterior mean and covariance accurately.

\subsection{Posterior Weights}
\label{sec:backward_posterior}
The true posterior distribution can be obtained from marginalizing over $\a$ according to
\begin{align}
	p_i(\vw) 
	&= \int_\NewR p(a,\vw|\SD_i) \dd a \\
	&= \int_\NewR \underbrace{p(\vw|a, \SD_i)}_\text{(I)} \cdot \underbrace{p(a|\SD_i)}_\text{(II)} \dd a~,
	\label{eq:backward_posterior}
\end{align}
which requires knowing two conditional distributions. The first one, indicated by (I), can be easily obtained by recalling from \Sec{sec:forward_output} that $\rvw$ and $\a$ are jointly Gaussian due to the linear mapping~\eqref{eq:problem_linear}. According to \cite{Rasmussen2006, Habil2015}, (I) can be written as
\begin{align}
	p(\vw|a, \SD_i) = \Gauss\hspace{-.4mm}\(\vw; \vmu_{i-1}^w + \vec l_i\cdot(a-\mu_a), \C_{i-1}^w - \vec l_i\cdot\vec\sigma_{wa}\T\)\quad~
	\label{eq:backward_conditionalw}
\end{align}
with gain vector $\vec l_i \DEF \nicefrac{\vec\sigma_{wa}}{\sigma_a^2}$. The mean $\vmu_{i-1}^w$ and covariance matrix $\C_{i-1}^w$ are given by the prior distribution, while $\mu_a$ and $\sigma_a^2$ are given by \eqref{eq:mua} and \eqref{eq:sigmaa}, respectively. For the covariance $\vec \sigma_{wa}$ it holds that
\begin{align}
	\vec \sigma_{wa} 
	&= \E\{(\vw - \vmu_{i-1}^w)\cdot (a - \mu_a)\} \\
	&= \E\hspace{-.5mm}\big\{(\vw - \vmu_{i-1}^w)\cdot (\vw - \vmu_{i-1}^w)\T\big\}\cdot \vx_i = \C_{i-1}^w\cdot \vx_i~.
	\label{eq:backward_sigmawa}
\end{align}
Thus, all ingredients of \eqref{eq:backward_conditionalw} are already available.

The conditional distribution (II) in \eqref{eq:backward_posterior} can be obtained from Bayes' rule according to
\begin{align}
	p(a|\SD_i) = \tfrac{1}{c} \cdot p(y_i|a)\cdot p(a|\vx_i, \SD_{i-1})
	\label{eq:backward_bayes}
\end{align}
with normalization constant $c = \int p(y_i|a)\cdot p(a|\vx_i, \SD_{i-1})\dd a$. Evaluating \eqref{eq:backward_bayes} in closed-form is not possible in general. However, assuming that $\y$ and $\a$ are jointly Gaussian yields a Gaussian approximation \cite{Habil2015}, i.e., $p(a|\SD_i) \approx \Gauss(a; \mu_i, \sigma_i^2)$ with mean and variance
\begin{align}
	\begin{split}
	\mu_i &= \mu_a + k_i\cdot(y_i-\mu_y) 
	\\
	\sigma_i^2 &= \sigma_a^2 - k_i\cdot \sigma_{ya}^2
	\end{split}
	\label{eq:backward_updateda}
\end{align}
respectively, with gain $k_i \DEF \nicefrac{\sigma_{ya}^2}{\sigma_y^2}$. This corresponds to the measurement update step of the famous Kalman filter \cite{Kalman1960}. All terms of the right-hand sides of \eqref{eq:backward_updateda} are known but $\sigma_{ya}^2$, which is given by 
\begin{align}
	\sigma_{ya}^2 
	&= \E\{(y-\mu_y)\cdot(a-\mu_a)\} \\
	&= \E\{a\cdot f(a)\} - \mu_y\cdot\mu_a~.
	\label{eq:backward_sigmaya}
\end{align}
So, it remains evaluating the expected value on the right-hand side of \eqref{eq:backward_sigmaya}.

\subsection{Sigmoid Case}
\label{sec:backward_sigmoid}
As it is the case for \eqref{eq:muy} and \eqref{eq:sigmay}, the sigmoid activation function hinders an analytical solution of the exptected value in \eqref{eq:backward_sigmaya}. Substituting the scaled probit function for the sigmoid function yields
\begin{align}
	\sigma_{ya}^2 
	&= \E\{a\cdot s(a)\} - \mu_y\cdot \mu_a \approx \E\{a\cdot \phi(\lambda\cdot a)\} - \underbrace{\mu_y\cdot \mu_a}_{\DEF \mu_{ya}} \\[-2ex]
	&= \int_\NewR a\cdot \phi(\lambda\cdot a) \cdot p_x(a) \dd a -\mu_{ya} \\
	&= \int_\NewR a\cdot \phi(\lambda\cdot a) \cdot \tfrac{1}{\sigma_a}\Gauss\hspace{-.5mm}\(\tfrac{a-\mu_a}{\sigma_a}; 0, 1\) \dd a -\mu_{ya} \\
	&\hspace{-.5mm}\stackrel{(g)}{=} \sigma_a\cdot\hspace{-1mm}\int_\NewR\hspace{-.5mm} z\cdot \phi(\lambda\hspace{-.5mm}\cdot\hspace{-.5mm}(\sigma_a\cdot z + \mu_a)) \cdot \Gauss(z; 0, 1) \dd z + \ldots \\
	&\hspace{5mm} \mu_a \cdot\hspace{-1mm} \underbrace{\int_\NewR\phi(\lambda\hspace{-.5mm}\cdot\hspace{-.5mm}(\sigma_a\cdot z + \mu_a)) \cdot \Gauss(z; 0, 1) \dd z}_{= \mu_y} -\,\mu_{ya} \\[-1ex]
	&= \sigma_a\cdot\hspace{-1mm}\int_\NewR z\cdot \phi(\lambda\cdot (\sigma_a\cdot z + \mu_a)) \cdot \Gauss(z; 0, 1) \dd z~,
	\label{eq:backward_sigmaya-sigmoid}
\end{align}
where $(g)$ follows from the change of variables $z = \tfrac{a-\mu_a}{\sigma_a}$. The integral in \eqref{eq:backward_sigmaya-sigmoid} has a closed-form expression (cf. \cite{Owen1980}, page 404, equation~10,011.3), such that $\sigma_{ya}^2$ can be closely approximated by means of
\begin{align}
	\sigma_{ya}^2 \approx \tfrac{\lambda\cdot \sigma_a^2}{t}\cdot \Gauss\hspace{-.4mm}\(\tfrac{\lambda\cdot \mu_a}{t}; 0, 1\)~.
	\label{eq:backward_sigmaya-sigmoid-final}
\end{align}

\subsection{Piece-wise Linear Case}
\label{sec:backward_pwl}
In contrast to the sigmoid case, pwl activation functions allow for a closed-form calculation of the expected value in~\eqref{eq:backward_sigmaya} according to
\begin{align}
	\sigma_{ya}^2 
	&= \E\{a\cdot f(a)\} - \mu_{ya} \\
	&= \int_\NewR a\cdot \max(\alpha\cdot a, \beta\cdot a)\cdot p_x(a) \dd a - \mu_{ya} \\
	&= \alpha\cdot\hspace{-.5mm}\int_{-\infty}^0\hspace{-.5mm} a^2\cdot p_x(a)\dd a + \beta \cdot\hspace{-.5mm}\int_0^\infty\hspace{-.5mm} a^2 \cdot p_x(a)\dd a - \mu_{ya} \\
	&= \alpha\cdot \E_2 +\, (\beta-\alpha)\cdot\(\E_2\cdot\,\phi\(\tfrac{\mu_a}{\sigma_a}\) + \mu_a\cdot p_a\) - \mu_{ya}~,
	\label{eq:backward_sigmaya-pwl}
\end{align}
which matches with \eqref{eq:sigmay_pwl} when replacing $\alpha^2$, $\beta^2$, $\mu_y^2$ with $\alpha$, $\beta$, $\mu_{ya}$, respectively.

\subsection{Summary}
\label{sec:backward_summary}
With the closed-form expressions of the covariance $\sigma_{ya}^2$ the conditional distribution in \eqref{eq:backward_updateda} is completely defined. It remains solving the marginalization in \eqref{eq:backward_posterior} to complete the update of the weights $\rvw$ given the $i$-th training instance $(\vx_i,y_i)$. As both (I) and (II) are Gaussian, calculating the product and solving the integral can be performed analytically exactly. In doing so, the true posterior $p_i(\vw)$ is approximated with the Gaussian $\Gauss(\vw; \vmu_i^w, \C_i^w)$ with mean and covariance given by
\begin{align}
	\begin{split}
	\vmu_i^w &= \vmu_{i-1}^w + \vec l_i\cdot(\mu_i - \mu_a)~, \\
  \C_i^w &= \C_{i-1}^w + \vec l_i\cdot(\sigma_i^2 - \sigma_a^2)\cdot\vec l_i\T~,
	\end{split}
	\label{eq:backward_weightupdate}
\end{align}
respectively, with gain $\vec l_i = \nicefrac{(\C_{i-1}^w\cdot\, \vx_i)}{\sigma_a^2}$ as in \eqref{eq:backward_conditionalw}. This completes the weight update. It is worth mentioning that the update equations in \eqref{eq:backward_weightupdate} coincide with the so-called \emph{Rauch-Tung-Striebel smoother} \cite{RTS1965}.

All calculations are summarized in \Alg{alg:forward} for the forward pass, i.e., inferring the predictive distribution given a test input $\vx$, and in \Alg{alg:backward} for the backward pass, i.e., training the proposed BP.

\begin{algorithm}[t]
\caption{Forward Pass for test input $\vx$}
\label{alg:forward}
\begin{algorithmic}[1]
\State Calculate mean $\mu_a$ via \eqref{eq:mua} and variance $\sigma_a^2$ via \eqref{eq:sigmaa}
\State \textbf{switch} activation function
\State \hspace{5.5mm}\emph{sigmoid:} Calculate mean $\mu_y$ and variance $\sigma_y^2$ of 
\Statex \hspace{19mm}output $\y$ according to \eqref{eq:muy_sigmoid} and \eqref{eq:sigmay_sigmoidapprox}
\State \hspace{5.5mm}\emph{pwl:} \hspace{6mm}Calculate mean $\mu_y$ and variance $\sigma_y^2$ of 
\Statex \hspace{19mm}output $\y$ according to \eqref{eq:my_pwl} and \eqref{eq:sigmay_pwl}
\State \textbf{end switch}
\State Return $\(\mu_y, \sigma_y^2, \mu_a, \sigma_a^2\)$
\end{algorithmic}
\end{algorithm}

In order to increase the robustness of learning, it is recommended to add a small positive term $\epsilon$ to the variance $\sigma_y^2$ in \eqref{eq:sigmay}. This corresponds to adding a zero-mean noise term $\v$ to~\eqref{eq:problem_activation} with variance $\epsilon$ according to
\begin{align}
	\y = f(a) + \v~,~\v\sim \Gauss(0, \epsilon)~.
	\label{eq:backward_noisy-activation}
\end{align}
This allows better capturing the noisy nature of the training data. Considering an additional noise term is common in Bayesian NN literature, e.g., \cite{Foong2019, Gal_ICML2016, Lobato_ICML2015}.

Learning a standard perceptron is inspired by so-called Hebbian learning \cite{Hebb2002}, where the weights are adapted whenever the output is not coinciding with a training instance according~to 
\begin{align}
	\vw_\mathrm{new} = \vw_\mathrm{old}  + \alpha\cdot \(y_i - f(\vx_i\T\cdot \vw_\mathrm{old})\)\cdot \vx_i~,
	\label{sec:backward_percepton}
\end{align}
which is known as the \emph{perceptron learning rule}, with $\alpha>0$ being the learning rate. 
The update of the weight's mean of the proposed BP follows a similar rule. By plugging the mean part of \eqref{eq:backward_updateda} in the mean update of \eqref{eq:backward_weightupdate} and by resolving the gain vector $\vec l_i$, the mean update becomes
\begin{align}
	\vmu_i^w = \vmu_{i-1}^w + \tfrac{k_i}{\sigma_a^2}\cdot \C_{i-1}^w\cdot (y_i - \mu_y)\cdot \vx_i~.
	\label{sec:backward_meanupdate}
\end{align}
By comparing \eqref{sec:backward_meanupdate} with \eqref{sec:backward_percepton} it can be seen that the term $\nicefrac{k_i}{\sigma_a^2}\cdot \C_{i-1}^w$ can be considered a matrix-valued learning rate. In contrast to the perceptron learning rule, the learning rate of the BP is neither constant nor identical for each weight. Even more important, thanks to the covariance matrix $\C_i^w$, updating an individual weight is influenced by all other weights if the weights are correlated. It is expected that this leads to a speed-up in training.

Like the perceptron learning rule, training the BP is gradient-free and can be performed sequentially. The latter allows applying the BP in learning tasks, where the training data is not available as a batch but becomes available over time. Both characteristics, i.e., being gradient-free and enabling sequential learning, are considered highly beneficial when employing the BP as a building block of Bayesian NN with multiple layers. Currently, Bayesian NN usually require the computationally expensive calculation of gradients in order to learn the network iteratively in a batch-wise manner. 

\begin{algorithm}[t]
\caption{Backward Pass for training the proposed BP with data $\SD$}
\label{alg:backward}
\begin{algorithmic}[1]
\State Initialize weight distribution with mean vector $\vmu_0^w$ and covariance matrix $\C_0^w$
\For{each training instance $(\vx_i, y_i) \in \SD$}
\State $(\mu_y, \sigma_y^2, \mu_a, \sigma_a^2) \leftarrow$ ForwardPass$(\vx_i)$ \hfill \Comment{\Alg{alg:forward}}
\State \textbf{switch} activation function
\State \hspace{5.5mm}\emph{sigmoid:} Calculate covariance $\sigma_{ya}^2$ via \eqref{eq:backward_sigmaya-sigmoid-final} 
\State \hspace{5.5mm}\emph{pwl:} \hspace{6mm}Calculate covariance $\sigma_{ya}^2$ via \eqref{eq:backward_sigmaya-pwl} 
\State \textbf{end switch}
\State Calculate mean $\mu_i$ and variance $\sigma_i^2$ according to \eqref{eq:backward_updateda}
\State Calculate gain vector $\vec l_i = \nicefrac{(\C_{i-1}^w\cdot\, \vx_i)}{\sigma_a^2}$
\State Update mean vector $\vmu_i^w$ and covariance matrix $\C_i^w$ 
\Statex \hspace{5.5mm}of the weights according to \eqref{eq:backward_weightupdate}
\EndFor
\State Return $\(\vmu^w, \C^w\)\leftarrow\big(\vmu_n^w, \C_n^w\big)$
\end{algorithmic}
\end{algorithm}

\section{Validation}
\label{sec:validation}
In this section, the proposed BP is validated in different learning tasks. At first binary classification is considered, where we compare the proposed moment approximations with the ground truth in \Sec{sec:validation_comparison} and where we demonstrate the on-line learning capabilities of the BP in \Sec{sec:validation_classification}. Then, in \Sec{sec:validation_regression} we employ on a nonlinear regression task to compare the proposed approach against gradient-based learning.

\subsection{Comparison with Ground Truth}
\label{sec:validation_comparison}
In order to quantify the quality of the proposed approximations a binary classification problem is considered, where we employ a perceptron with sigmoid activation function. Here, the focus is on the conditional distribution $p(a|\SD_i)$ in \eqref{eq:backward_posterior}, which is approximated by means of a Gaussian distribution. Focusing on this part of the BP is beneficial for two reasons: (i) the considered quantities are one-dimensional, which allows for numerical integration to obtain the ground truth and (ii) it covers all approximations, i.e., approximating the mean $\mu_y$, variance $\sigma_y^2$ and covariance $\sigma_{ya}^2$ via \eqref{eq:muy_sigmoid}, \eqref{eq:sigmay_sigmoidapprox}, and \eqref{eq:backward_sigmaya-sigmoid-final}, respectively, due to the employed sigmoid activation and approximating $p(a|\SD_i)$ by a Gaussian with mean and variance according to \eqref{eq:backward_updateda}.

\begin{figure}%
\centering
\begin{tikzpicture}
	\node at (0,0) {\includegraphics[]{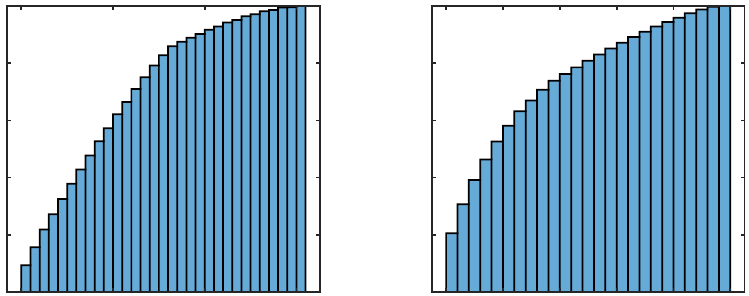}};
	
	\node at (-3.57,-1.7) {0};
	\node at (-2.67,-1.7) {0.1};
	\node at (-1.72,-1.7) {0.2};
	\node at (-.8,-1.7) {0.3};
	\node at (-2.1,-2.1) {abs. error $\rightarrow$};
	
	\node at (-3.9,-1.45) {0};
	\node at (-4.05,-.9) {0.2};
	\node at (-4.05,-.3) {0.4};
	\node at (-4.05,.27) {0.6};
	\node at (-4.05,.85) {0.8};
	\node at (-3.9,1.45) {1};
	\node at (-4.6,0) {\rotatebox{90}{cum. probability $\rightarrow$}};
	\node at (-4.4,-2.1) {(a)};

	\node at (.73,-1.7) {0};
	\node at (1.3,-1.7) {0.1};
	\node at (1.9,-1.7) {0.2};
	\node at (2.45,-1.7) {0.3};
	\node at (3.05,-1.7) {0.4};
	\node at (3.63,-1.7) {0.5};
	\node at (2.1,-2.1) {abs. error $\rightarrow$};
	
	\node at (.4,-1.45) {0};
	\node at (.25,-.9) {0.2};
	\node at (.25,-.3) {0.4};
	\node at (.25,.27) {0.6};
	\node at (.25,.85) {0.8};
	\node at (.4,1.45) {1};
	\node at (-.3,0) {\rotatebox{90}{cum. probability $\rightarrow$}};
	\node at (-.1,-2.1) {(b)};
\end{tikzpicture}
\caption{Cumulative distribution of the errors in posterior mean (a) and variance (b) between ground truth and proposed solution.}%
\label{fig:comparison}%
\end{figure}
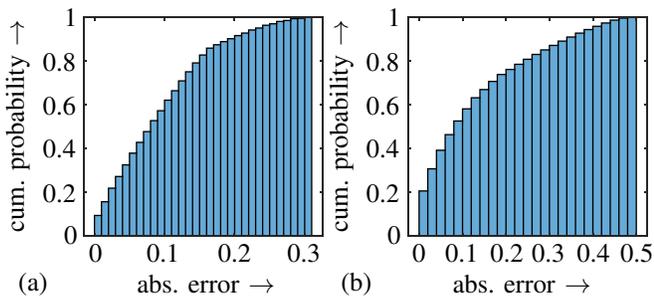

To compare the results of the BP with the ground truth generated by numerically solving \eqref{eq:backward_bayes}, we vary the values of the mean $\mu_a$ and variance $\sigma_a^2$ over a wide range. More precisely, $\mu_a$ takes values from the set $\{-3, -2.9, -2.8, \ldots, 3\}$ while $\sigma_a^2 \in \{0, 0.2, 0.4, \ldots, 2\}$. The output $y \in \{0,1\}$ is determined based on the mean $\mu_a$ and the Heaviside step function according to
\begin{align}
	y = \begin{cases}
		1 &,\ \mu_a > 0 \\
		0 &,\ \text{otherwise}
	\end{cases}~.
\end{align}

In \Fig{fig:comparison} the cumulative distributions of the absolute errors between the true mean/variance and the approximate mean/variance given by \eqref{eq:backward_updateda} are depicted. It can be seen, that the approximation is very close to the ground truth over all combinations of values for $\mu_a$ and $\sigma_a^2$. Especially in case of the mean the absolute error is less than $0.2$ in more than $80\%$ of the cases. Accordingly, the mean absolute error (mae) for the mean $\mu_a$ is $0.0952 \pm 0.0708$. In case of the variance $\sigma_a^2$ the mae is slightly higher with $0.1349 \pm 0.1291$. These close approximations are obtained with a runtime being two orders of magnitude smaller than the ground truth calculations. This significant difference in runtime is important when scaling the BP up to a deep Bayesian NN consisting of hundreds or thousands of perceptrons/neurons.

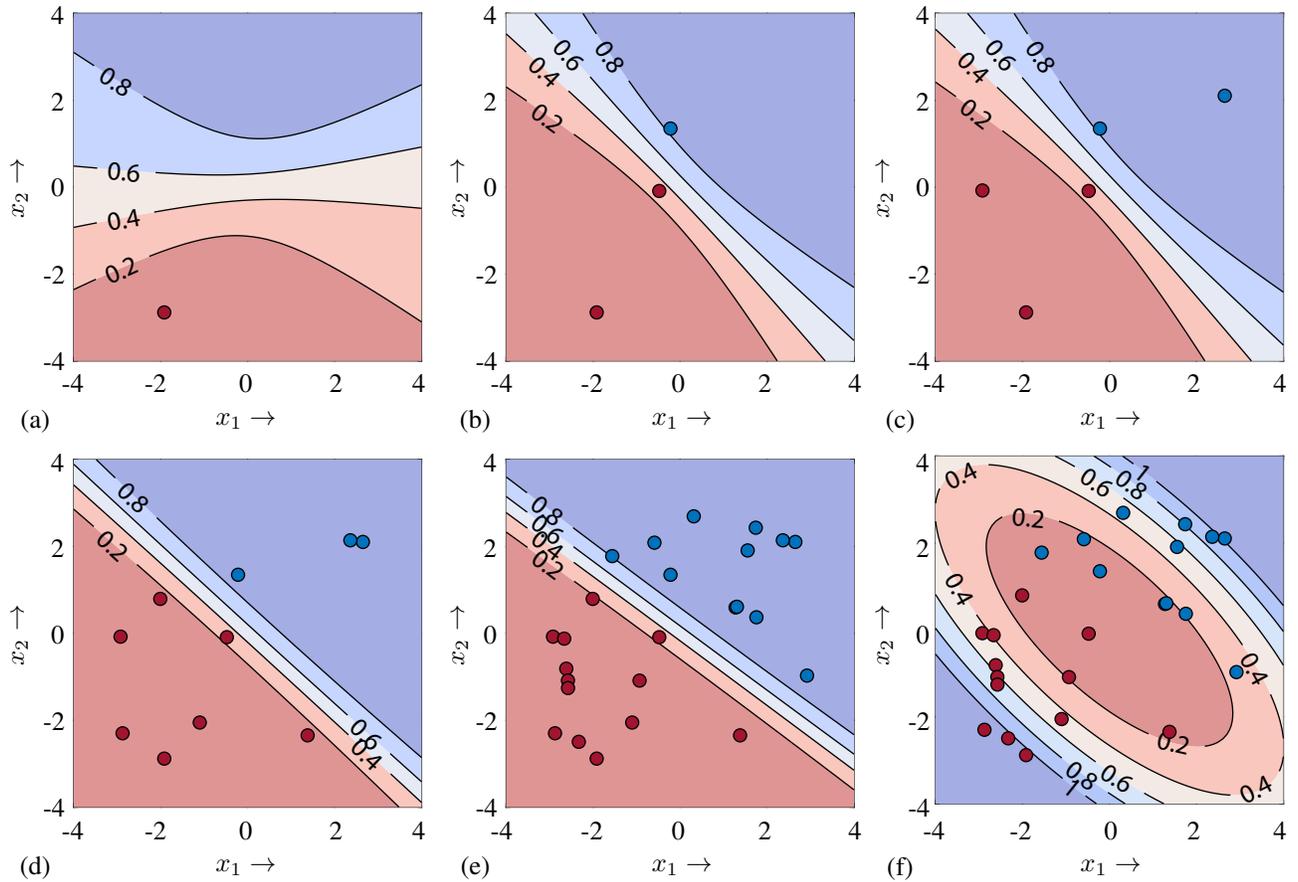
\begin{figure*}[t]%
\centering
\begin{tikzpicture}
	\node at (0,0) {\includegraphics[]{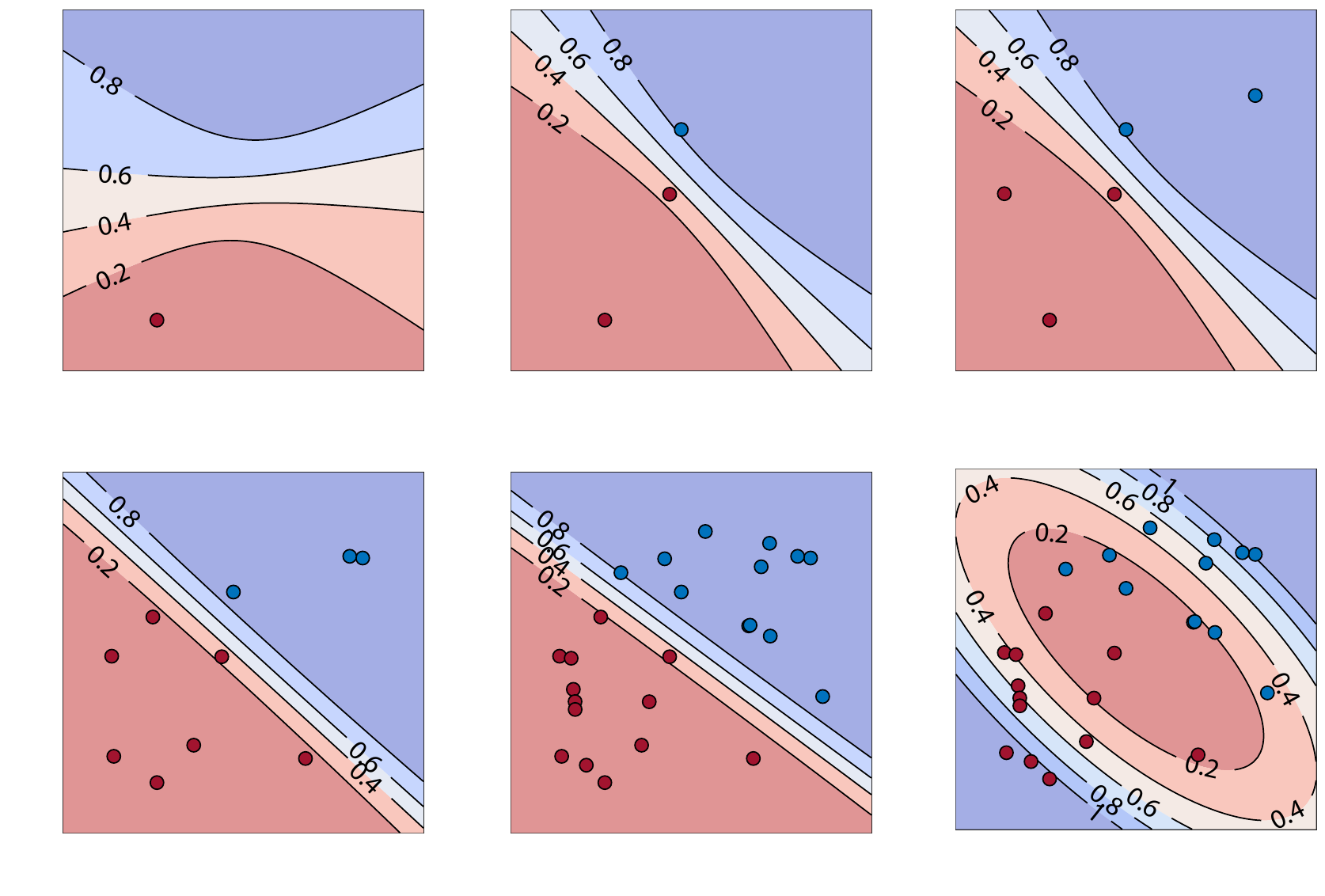}};
		
	\node at (-7.95,1) {-4};
	\node at (-7.95,2.15) {-2};
	\node at (-7.9,3.3) {0};
	\node at (-7.9,4.45) {2};
	\node at (-7.9,5.6) {4};
	\node at (-8.4,3.3) {\rotatebox{90}{$x_2 \rightarrow$}};
	\node at (-7.7,.7) {-4};
	\node at (-6.6,.7) {-2};
	\node at (-5.4,.7) {0};
	\node at (-4.2,.7) {2};
	\node at (-3.1,.7) {4};
	\node at (-5.4,.2) {$x_1 \rightarrow$};
	\node at (-8.2,.2) {(a)};
	
	\node at (-2.2,1) {-4};
	\node at (-2.2,2.15) {-2};
	\node at (-2.15,3.3) {0};
	\node at (-2.15,4.45) {2};
	\node at (-2.15,5.6) {4};
	\node at (-2.55,3.3) {\rotatebox{90}{$x_2 \rightarrow$}};
	\node at (-2,.7) {-4};
	\node at (-.8,.7) {-2};
	\node at (.35,.7) {0};
	\node at (1.5,.7) {2};
	\node at (2.7,.7) {4};
	\node at (.4,.2) {$x_1 \rightarrow$};
	\node at (-2.35,.2) {(b)};
	
	\node at (3.55,1) {-4};
	\node at (3.55,2.15) {-2};
	\node at (3.6,3.3) {0};
	\node at (3.6,4.45) {2};
	\node at (3.6,5.6) {4};
	\node at (3.1,3.3) {\rotatebox{90}{$x_2 \rightarrow$}};
	\node at (3.75,.7) {-4};
	\node at (4.9,.7) {-2};
	\node at (6.1,.7) {0};
	\node at (7.2,.7) {2};
	\node at (8.35,.7) {4};
	\node at (6.1,.2) {$x_1 \rightarrow$};
	\node at (3.3,.2) {(c)};

	\node at (-7.95,-4.95) {-4};
	\node at (-7.95,-3.8) {-2};
	\node at (-7.9,-2.65) {0};
	\node at (-7.9,-1.5) {2};
	\node at (-7.9,-.35) {4};
	\node at (-8.4,-2.65) {\rotatebox{90}{$x_2 \rightarrow$}};
	\node at (-7.7,-5.25) {-4};
	\node at (-6.6,-5.25) {-2};
	\node at (-5.4,-5.25) {0};
	\node at (-4.2,-5.25) {2};
	\node at (-3.1,-5.25) {4};
	\node at (-5.4,-5.75) {$x_1 \rightarrow$};
	\node at (-8.2,-5.75) {(d)};
	
	\node at (-2.2,-4.95) {-4};
	\node at (-2.2,-3.8) {-2};
	\node at (-2.15,-2.65) {0};
	\node at (-2.15,-1.5) {2};
	\node at (-2.15,-.35) {4};
	\node at (-2.55,-2.65) {\rotatebox{90}{$x_2 \rightarrow$}};
	\node at (-2,-5.25) {-4};
	\node at (-.8,-5.25) {-2};
	\node at (.35,-5.25) {0};
	\node at (1.5,-5.25) {2};
	\node at (2.7,-5.25) {4};
	\node at (.4,-5.75) {$x_1 \rightarrow$};
	\node at (-2.35,-5.75) {(e)};
	
	\node at (3.55,-4.95) {-4};
	\node at (3.55,-3.8) {-2};
	\node at (3.6,-2.65) {0};
	\node at (3.6,-1.5) {2};
	\node at (3.6,-.35) {4};
	\node at (3.1,-2.65) {\rotatebox{90}{$x_2 \rightarrow$}};
	\node at (3.75,-5.25) {-4};
	\node at (4.9,-5.25) {-2};
	\node at (6.1,-5.25) {0};
	\node at (7.2,-5.25) {2};
	\node at (8.35,-5.25) {4};
	\node at (6.1,-5.75) {$x_1 \rightarrow$};
	\node at (3.3,-5.752) {(f)};
\end{tikzpicture}
\caption{Evolution of the predicted mean $\mu_y = \text{Prob}(y=1|\vx, \SD)$ for (a) one, (b) three, (c) five, (d) ten, and (e) 25 data instances. The variance $\sigma_a^2$ is shown in (f). It is important to note that the variance is not limited to one. It continues growing but higher values are not plotted in different colors in order to keep the visualization simple. Red dots indicate data instances belonging to class $y=0$ and blue dots belong to class $y=1$.}
\label{fig:classification}%
\end{figure*}

\subsection{Linear Binary Classification}
\label{sec:validation_classification}
A further binary classification problem is considered in the following, where data is generated uniformly at random over the two-dimensional area $[-3, 3] \times [-3, 3]$. The data points $\vx\in \NewR^2$ are assigned to one of the two classes according to
\begin{align}
	y = \begin{cases}
	1 &,\ [1\ 1]\cdot \vx > 0 \\
	0 &,\ \text{otherwise}
	\end{cases}~.
	\label{eq:validation_decision-boundary}
\end{align}
In total $n=25$ data instances are generated and used for sequentially training a BP without bias term. The weight distribution is initialized with $\mu_0^w = [-1\ 0]\T$ and $\C_0^w = \I_2$ with $\I_2$ being the $2\times 2$ identity matrix. In \Fig{fig:classification}(a)--(e) the evolution of the predictive mean $\mu_y$ is depicted. It can been seen that the initially rather high indifference becomes continuously sharper and the perceptron is able to correctly learn the decision boundary being defined by \eqref{eq:validation_decision-boundary}. These results are representative for $50$ conducted random trials. 

\Fig{fig:classification}(f) shows the finale variance $\sigma_a^2$ after processing all data instances. This gives a good indication about the uncertainty quantification of the BP. In areas with many training data, the variance is low (red), while in areas with low data density or even no data, the uncertainty is growing (blue).

\begin{figure*}[t]%
\centering
\begin{tikzpicture}
	\node at (0,0) {\includegraphics[]{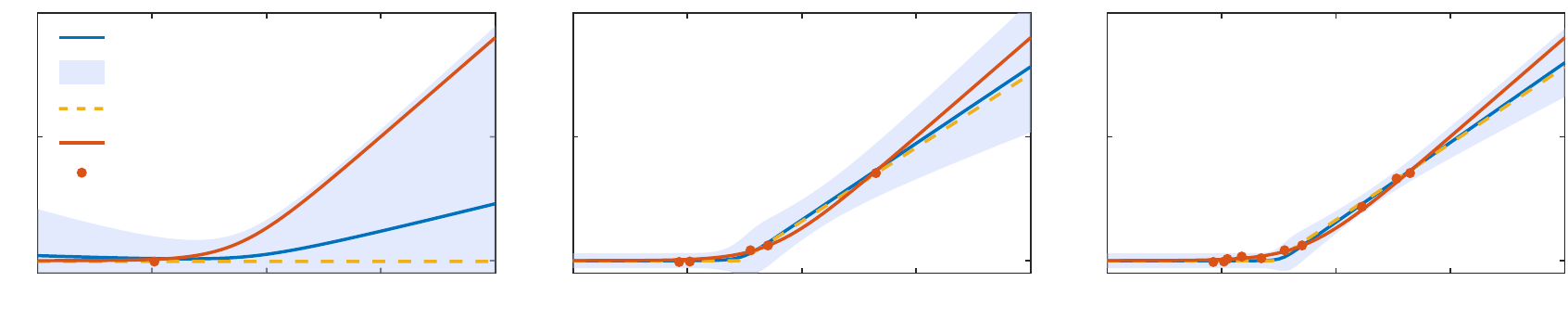}};
	
	\node at (-8.35,-1.1) {0};
	\node at (-8.35,.25) {5};
	\node at (-8.45,1.57) {10};
	\node at (-8.8, .25) {\rotatebox{90}{$y \rightarrow$}};
	
	\node at (-8.2,-1.5) {-4};
	\node at (-7.0,-1.5) {-2};
	\node at (-5.7,-1.5) {0};
	\node at (-4.4,-1.5) {2};
	\node at (-3.2,-1.5) {4};
	\node at (-5.7,-1.9) {$x \rightarrow$};
	\node at (-8.6,-1.9) {(a)};

	\node at (-2.48,-1.1) {0};
	\node at (-2.48,.25) {5};
	\node at (-2.58,1.57) {10};
	\node at (-2.9, .25) {\rotatebox{90}{$y \rightarrow$}};
	
	\node at (-2.35,-1.5) {-4};
	\node at (-1.1,-1.5) {-2};
	\node at (0.2,-1.5) {0};
	\node at (1.45,-1.5) {2};
	\node at (2.7,-1.5) {4};
	\node at (0.2,-1.9) {$x \rightarrow$};
	\node at (-2.7,-1.9) {(b)};

	\node at (3.37,-1.1) {0};
	\node at (3.37,.25) {5};
	\node at (3.27,1.57) {10};
	\node at (2.97, .25) {\rotatebox{90}{$y \rightarrow$}};
	
	\node at (3.5,-1.5) {-4};
	\node at (4.75,-1.5) {-2};
	\node at (6.05,-1.5) {0};
	\node at (7.3,-1.5) {2};
	\node at (8.54,-1.5) {4};
	\node at (6.05,-1.9) {$x \rightarrow$};
	\node at (3.15,-1.9) {(c)};
	
	\node at (-7.1,1.35) {\small BP};
	\node at (-6.72,.95) {\small BP (3-$\sigma$)};
	\node at (-6.75,.59) {\small Gradient};
	\node at (-7.0,.22) {\small True};
	\node at (-6.98,-.15) {\small Data};
\end{tikzpicture}
\caption{Sequentially learning the softplus function \eqref{eq:validation_softplus} (red) with (a) one, (b) five, and (c) ten data instances.}%
\label{fig:validation_softplus_example}%
\vspace{-2mm}
\end{figure*}
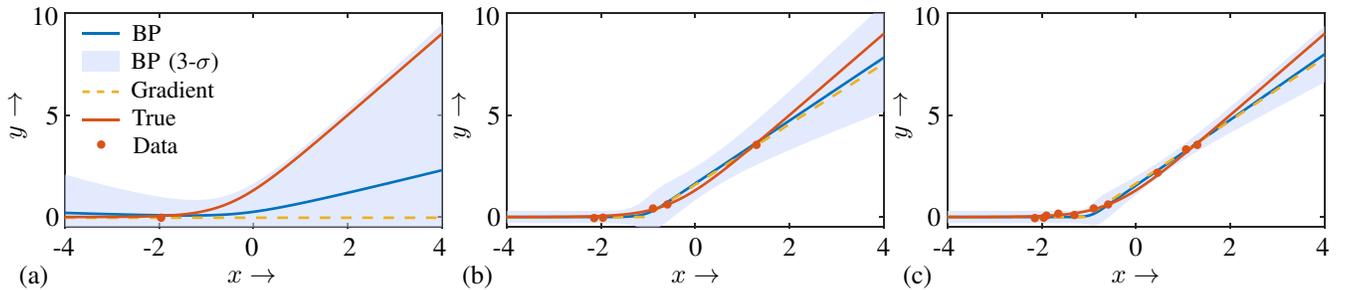

\subsection{Nonlinear Regression}
\label{sec:validation_regression}
Perceptrons are used for linear classification problems only, but depending on the activation function used, also simple regression problems can be tackled. Here, we consider a regression problem where the data is generated by means of a noisy \emph{softplus} function
\begin{align}
	\y = \log\(1 + \mathrm{e}^{\gamma\cdot x + \delta}\) + \v~,~\v\sim\Gauss(0, 0.01)~.
	\label{eq:validation_softplus}
\end{align}
The softplus function with parameters $\gamma = 1$ and $\delta=0$ is a smooth approximation of the ReLU function. Hence, we employ a BP with ReLU activation. To make the problem a bit more challenging, the parameters of the softplus function are set to be $\gamma = 2$ and $\delta=1$, so that there is a misfit between softplus and ReLU that needs to be compensated by learning appropriate weights. The weights of the BP are initialized with mean $\mu_0^w = [0\ 0]\T$ and covariance matrix $\C_0^w = \I_2$. For comparison, another (deterministic) perceptron is employed, where the weights are learned via error backpropagation, i.e., gradient descent. 

In total $50$ random trials are conducted, where for each trial a dataset of size $n=20$ training data instances is generated randomly by uniformly sampling $x$-values from the interval $[-4, 2]$. In \Fig{fig:validation_softplus_example} an exemplary training trial is depicted. It can be seen how both perceptrons approach the true function \eqref{eq:validation_softplus} with an increasing number of training instances. In contrast to the standard perceptron, the proposed BP provides an uncertainty quantification in addition by means of the predictive variance $\sigma_y^2$. In areas where there are no data, the predictions of the BP are less certain being indicated by the wider uncertainty 3-$\sigma$ band.

To quantify the accuracy, $40$ test instances are generated randomly over the interval $x \in [-4, 4]$. The averaged root-mean squarred error (rmse) for these test instances is depicted in \Fig{fig:validation_softplus}. Independent of the size of the processed training data, the proposed BP outperforms classical gradient-based training. Particularly with very small data, the performance of BP is less volatile being indicated with the tighter error bars. This better performance comes in addition with a significantly lower computational burden, which is measured by means of the runtime.

\section{Conclusions and Future Work}
\label{sec:conclusions}
In this paper, the Bayesian perceptron is introduced, which is a probabilistic extension of the classical perceptron. It allows calculating a probability distribution as an output and thus, gives an indication about the certainty of the prediction. The parameters of the Bayesian perceptron, i.e., its weights, are learned by means of Bayesian inference in closed form without the need of iterative gradient descent. Furthermore, learning can be performed sequentially, which makes the approach suitable for on-line learning and real-time applications. 

The proposed method is intended as the core building block of a new type of Bayesian NN training. Future work is devoted to extending the introduced Bayesian learning for a single perceptron to deep Bayesian neural networks.

\section*{Acknowledgments}
The author would like to thank Philipp Wagner for valuable comments and fruitful discussions.


\definecolor{matblue}{rgb}{0,0.45,0.74}
\definecolor{matred}{rgb}{.85,0.32,0.10}
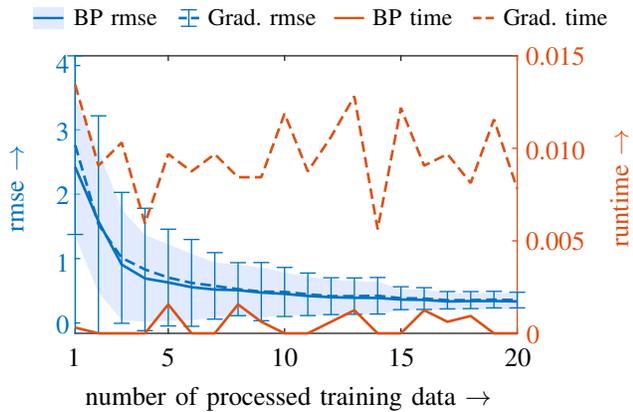
\begin{figure}[tb]%
\centering
\begin{tikzpicture}
	\node at (0,0) {\includegraphics[]{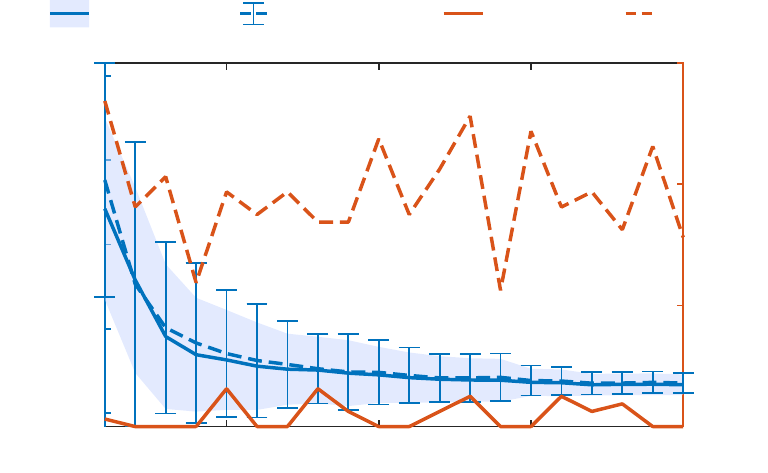}};
	
	\node[matblue] at (-3.0,-1.82) {0};
	\node[matblue] at (-3,-1.0) {1};
	\node[matblue] at (-3,-.1) {2};
	\node[matblue] at (-3,.75) {3};
	\node[matblue] at (-3,1.65) {4};
	\node[matblue] at (-3.6, 0) {\rotatebox{90}{rmse $\rightarrow$}};

	\node[matred] at (3.25,-1.95) {0};
	\node[matred] at (3.55,-.7) {0.005};
	\node[matred] at (3.55,.5) {0.010};
	\node[matred] at (3.55,1.75) {0.015};
	\node[matred] at (4.4, 0) {\rotatebox{90}{runtime $\rightarrow$}};
	
	\node at (-2.85,-2.25) {1};
	\node at (-1.6,-2.25) {5};
	\node at (-.05,-2.25) {10};
	\node at (1.5,-2.25) {15};
	\node at (3.05,-2.25) {20};
	\node at (0,-2.8) {number of processed training data $\rightarrow$};
	
	\node at (-2.3,2.27) {\small BP rmse};
	\node at (-.35,2.27) {\small Grad. rmse};
	\node at (1.65,2.27) {\small BP time};
	\node at (3.55,2.27) {\small Grad. time};
\end{tikzpicture}
\caption{Comparison of BP (solid) with gradient-based learning (dashed) w.r.t. rms error (blue) and runtime (red).}%
\label{fig:validation_softplus}%
\end{figure}

\balance

\end{document}